\documentclass[runningheads]{llncs}
\usepackage[T1]{fontenc}
\usepackage{multirow}
\usepackage{graphicx}
\usepackage{algorithm,algorithmic}
\usepackage[dvipsnames,table]{xcolor}
\usepackage{multirow}
\usepackage{amssymb,amsmath}
\usepackage{graphicx}
\usepackage[colorlinks]{hyperref}
\usepackage{booktabs}
\begin{document}
%
\title{APPLE: Adversarial Privacy-aware Perturbations on Latent Embedding for Unfairness Mitigation}
\titlerunning{APPLE}
%

\author{
            Zikang Xu\inst{1,2},
            Fenghe Tang\inst{1,2},
            Quan Quan\inst{3},
            Qingsong Yao\inst{3},
            S. Kevin Zhou\inst{1,2,3}
        }
\institute{
            School of Biomedical Engineering, Division of Life Sciences and Medicine, University of Science and Technology of China, Hefei, Anhui, 230026, P.R.China \and
            Suzhou Institute for Advanced Research, University of Science and Technology of China, Suzhou, Jiangsu, 215123, P.R.China \and
            Key Lab of Intelligent Information Processing of Chinese Academy of Sciences (CAS), Institute of Computing Technology, CAS, Beijing, 100086, P.R.China
            \email{skevinzhou@ustc.edu.cn}\\
}

\authorrunning{Xu~\textit{et al.}}

\maketitle              
\begin{abstract}
    Ensuring fairness in deep-learning-based segmentors is crucial for health equity.
    Much effort has been dedicated to mitigating unfairness in the training datasets or procedures.
    However, with the increasing prevalence of foundation models in medical image analysis, it is hard to train fair models from scratch while preserving utility.
    In this paper, we propose a novel method, Adversarial Privacy-aware Perturbations on Latent Embedding (APPLE), that can improve the fairness of deployed segmentors by introducing a small latent feature perturber without updating the weights of the original model.
    By adding perturbation to the latent vector, APPLE decorates the latent vector of segmentors such that no fairness-related features can be passed to the decoder of the segmentors while preserving the architecture and parameters of the segmentor.
    Experiments on two segmentation datasets and five segmentors (three U-Net-like and two SAM-like) illustrate the effectiveness of our proposed method compared to several unfairness mitigation methods.

\end{abstract}
\section{Introduction}

In recent years, deep-learning (DL) based algorithms have shown powerful performances in various medical applications, including classification, segmentation, detection, and low-level tasks such as reconstruction and denoising~\cite{zhou2021review}.
Apart from making efforts to improve model utilities on these tasks, more and more studies recognize an important issue about the DL models, i.e., they may have \textbf{unfair} performances on subgroups with different sensitive attributes~\cite{seyyed2021underdiagnosis}.
This is a vital problem that cannot be ignored as model unfairness hurts people's rights to be treated equally, reduces the reliability and trustworthiness of DL models, and does great harm to privacy-preserving and health equity~\cite{xu2023fairness}.

Originally, researchers only focused on unfairness in classification tasks, adopting general fairness metrics introduced in the machine learning area including Demographic Parity (DP)~\cite{dwork2012fairness}, Equality of Odds (EqOdd)~\cite{hardt2016equality}, etc.
However, Puyol-Ant\'on~\textit{et al.}~\cite{puyol2021fairness} find that unfairness also exists in segmentation tasks, which piques people's interest in mitigating unfairness in other medical applications besides classification, and protecting the patients' \textbf{privacy}, especially the sensitive ones including \textit{age}, \textit{sex}, \textit{race}, etc.

On the other hand, with the recent development in large-scale medical foundation models, such as MedSAM~\cite{MedSAM} and Slide-SAM~\cite{quan2024slide}, an increasing number of studies try to adopt the outstanding utility of pre-trained models to the target task.
However, although these models show competitive few-shot or zero-shot utilities on the target tasks due to their huge amount of train data, some studies witness that these models might inherit the biases from training data and perform unfairly in downstream tasks~\cite{booth2021bias}, which is unacceptable for ensuring health equity. 
As re-training or fine-tuning these models requires heavy computing resources and a long time, there is a need to come up with methodologies that can ensure fairness while preserving the utility of the pre-trained models. 

In this paper, we consider a setting where the baseline segmentors are hard to re-train, which limits the degrees of freedom and brings difficulties for unfairness mitigation.
Therefore, we need to find algorithms that can improve fairness \textbf{without changing the parameters of the base model}.
Inspired by studies about perturbation synthesis using generative adversarial networks (GAN)~\cite{xiao2018generating}, we propose a novel method, denoted as \textbf{Adversarial Privacy-aware Perturbations on Latent Embedding (APPLE)} to mitigate unfairness in pre-trained medical segmentors by manipulating the extracted latent Embedding with privacy-aware perturbations.
Specifically, APPLE consists of a generator that perturbs the latent embedding of the fixed segmentation encoder, trying to hide the sensitive attribute of the input image, and a discriminator aims to distinguish the sensitive attribute from the perturbed latent Embedding.
Compared to methods that directly perturb the input image~\cite{wang2022fairness}, the perturbation in the latent space is more effective as the features contain higher semantic information and are more separable in the deeper feature space.
Besides, as a model-agnostic algorithm, APPLE can be applied in almost any segmentors, as long as the segmentor can be split into a feature encoder and a prediction decoder. 

Our contributions are four folds:
\begin{enumerate}
    \item We implement the first attempt at unfairness mitigation in medical segmentation where the parameters are fixed, which is more practical for medical applications;
    \item We propose APPLE, which improves model fairness by perturbing the latent Embedding using a GAN. By adjusting the weight factor $\beta$ of the loss function, APPLE can control the extent of unfairness mitigation;
    \item Extensive experiments on two medical segmentation datasets prove that APPLE can improve fairness in deployed segmentors effectively;
    \item Experiments integrating APPLE with large-scale foundation segmentation models including SAM and MedSAM prove the promising utility of APPLE for unfairness mitigation on foundation models. 
\end{enumerate}

\section{Related Work}

\noindent{\textbf{Fairness in Medical Image Analysis (MedIA)}}
Various studies about fairness in medical image analysis have been conducted and these researches vary in image modalities (X-ray~\cite{seyyed2021underdiagnosis}, MRI~\cite{lee_investigation_2023}, Dermatology~\cite{xu2023fairadabn}) and body parts (Brain~\cite{petersen2022feature}, Chest~\cite{pmlr-v174-zhang22a}, Retinal~\cite{burlina2021addressing}).
Some address fairness issues in MedIA~\cite{seyyed2021underdiagnosis}, and others develop unfairness mitigation methods, which are categorized into pre-processing~\cite{yao2022improving}, in-processing~\cite{deng2023fairness}, and post-processing~\cite{wu2022fairprune}.
There are also few studies addressing fairness in other tasks besides classification.
For example, Puyol-Ant\'on~\textit{et al.} conduct continuous experiments evaluating fairness in heart segmentation tasks~\cite{puyol2021fairness}. Ioannou~\textit{et al.}~\cite{ioannou2022study} study biases and unfairness in brain MRI. And Du~\textit{et al.}~\cite{du2023unveiling}, unveil fairness in MRI reconstruction tasks.

\noindent{\textbf{Improving Fairness via Data }}
Several attempts try to improve fairness by data manipulation.
This can be done by generating new samples with the same target label while opposing sensitive attribute~\cite{pakzad2022circle}, removing `culprit' background artifacts from the background~\cite{bissoto2020debiasing}, or generating sketches of the origin dataset by language-guided models to remove sex and age information from the image~\cite{yao2022improving}.
However, these methods require re-training or fine-tuning the pre-trained models, which is time-consuming and not applicable in our settings.

\section{APPLE: Adversarial Privacy-aware Perturbations on Latent Embedding}
\subsection{Problem Definition}

Supposing a segmentation dataset $D=\{(x_i, y_i, a_i)\}_{i=1}^{N}$, where $x_i$ is the input image from a specific patient, $y_i$ is the target segmentation mask, and $a_i$ is the sensitive attributes (e.g. sex or age).
The dataset is split into $K$ mutually exclusive subgroups by the privacy metadata of the patient, also known as the sensitive attributes, $a\in R^K$. i.e., $D = D_1 \cup D_2 \ldots \cup D_K$ and $D_i \cap D_j = \emptyset (\forall i, j \in \{1, 2, \ldots, K\}, i \neq j)$.
Fair segmentation requires the minimum segmentation utility discrepancies among the $K$ groups.

\subsection{Fair Perturbation}

Generally, a segmentation network $\phi$ can be split into the encoder part $E_s$ and the decoder part $D_s$, i.e. $\phi(x) = D_s(E_s(x))$. 
Note that in our settings, the parameters of $E_s$ and $D_s$ are  \textbf{fixed}, which limits the degree of freedom for unfairness mitigation.

\begin{figure}[t]
    \centering
    \includegraphics[width=0.9\textwidth]{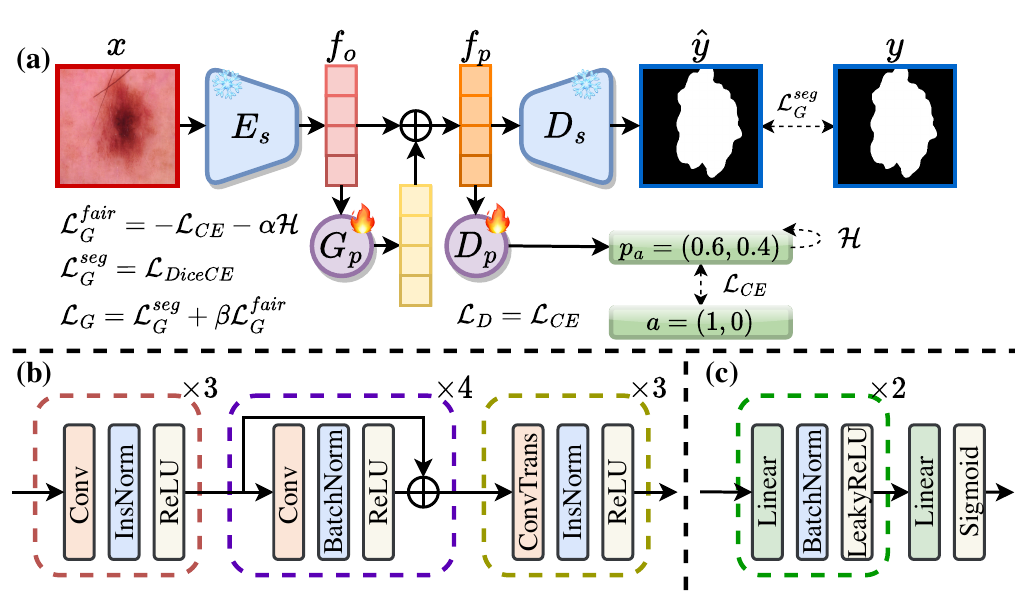}
    \caption{Flowchart of APPLE.  (a) Overall pipeline. $E_s$ and $D_s$ are frozen, only $G_P$ an $D_p$ are trainable; (b) Architecture of $G_p$; (c) Architecture of $D_p$.}
    \label{fig:pipeline}
\end{figure}

As shown in~\cite{kim2019learning}, the prediction of a model is fairer if less sensitive attribute-related information is encoded into the latent feature Embedding. This mainly results from the less usage of spurious relationships between the confounding attributes and the target task.
Based on the above motivation, the proposed APPLE tries to mitigate unfairness by manipulating the latent Embedding, $f_o = E_s(x)$, which decorates the $f_o$ such that the sensitive attributes cannot be recognized from the perturbed Embedding using generative adversarial networks. The pipeline of APPLE is shown in Fig.~\ref{fig:pipeline}.

Specifically, for an input image $x$, we extract the latent embedding $f_o$ using $E_s$. After that, a privacy-aware perturbation generator $G_p$ is used to learn \textit{fair perturbations}, i.e. $f_p = f_o + G_p(f_o)$.
Then $f_p$ is passed to the decoder $D_s$ to predict the modified segmentation mask, $\hat{y}$, given by Equ.~\ref{decoder}.

\begin{equation}
    \hat{y} = D_s(f_p) = D_s(f_o + G_p(f_o)) = D_s(E_s(x) + G_p(E_s(x)).
    \label{decoder}
\end{equation}

\subsection{Architecture of $G_p$ and $D_p$}

The $G_p$ consists of a 3-layer encoder with channels of (32, 64, 128), a 4-layer bottleneck, and a 3-layer decoder with channels of (64, 32, $N_{embedding-channels}$).
The $D_p$ is composed of 3 Linear-BatchNorm blocks, and the output dim is set to the number of classes of sensitive attributes.
The detailed architectures of $G_p$ and $D_p$ are shown in Fig.~\ref{fig:pipeline}.

\subsection{Loss Functions}
The discriminator $D_p$ tries to distinguish the sensitive attribute from the perturbed latent embedding $f_p$, which can be optimized by Equ.~\ref{discriminator}.

\begin{equation}
    \mathcal{L}_{D} = \mathcal{L}_{CE}(D_p(f_p), a).
    \label{discriminator}
\end{equation}

The loss function of the perturbation generator $G_p$ consists of two parts: the segmentation utility preserving part $\mathcal{L}_G^{seg}$ and the fairness constraining part $\mathcal{L}_G^{fair}$.
$\mathcal{L}_G^{seg}$ is defined as the Dice-CE Loss between the predicted mask and the ground truth mask:

\begin{equation}
    \mathcal{L}_G^{seg} = \frac{1}{2}(\mathcal{L}_{CE}(y, \hat{y}) + \mathcal{L}_{Dice}(y, \hat{y})).
\end{equation}

The fairness constraints part aims to generate perturbations that confuse the distinction of the sensitive attribute, which is controlled by the negative cross-entropy loss of sensitive attribute prediction. Besides, a regularization term on the entropy of $D_p(f_p)$ is added to avoid space collapse. Therefore, $\mathcal{L}_G^{fair}$ is given by the following Equation:

\begin{equation}
    \mathcal{L}_G^{fair} = -\mathcal{L}_{CE}(D_p(f_p), a) - \alpha \mathcal{H}(D_p(f_p)),
\end{equation}
where $\mathcal{H}$ is the entropy.
Therefore, the overall loss of $G_p$ is defined as Equ.~\ref{loss_g}: 

\begin{equation}
    \mathcal{L}_G = \mathcal{L}_G^{seg} + \beta\mathcal{L}_{G}^{fair},
    \label{loss_g}
\end{equation}
where $\beta$ is the weighted factor that controls the degree of requirement on fairness.

Following adversarial training, our proposed APPLE is optimized by the min-max game using Algorithm~\ref{pseduocode}.

In the test stage, the input image is firstly sent to the segmentation encoder $E_s$ to get latent embedding $f_o$. Then the perturbed embedding $f_p = f_o + E_p(f_o)$ is passed to the segmentation decoder $D_s$ to generate the final prediction mask $\hat{y}$. 

\begin{algorithm}[t]
    \caption{\textbf{Training Procedure of APPLE}}
    \label{pseduocode}
    \begin{algorithmic}
        \REQUIRE{Encoder $E_s$ and Decoder $D_s$ of fixed segmentor, input $x$, segmentation mask $y$, sensitive attribute $a$, weighted factor $\alpha, \beta$, number of epoch $N_{epoch}$}
        \ENSURE{Perturbation Generator $G_p$ and Discriminator $D_p$}

        \FOR{$i = 1: N_{epoch}$}
            \STATE Get the original latent embedding: $f_o \leftarrow E_s(x)$.
            \STATE Get the perturbed latent embedding: $f_p \leftarrow f_o + G_p(f_o)$.
            \STATE Get the predicted segmentation mask: $\hat{y} \leftarrow D_s(f_p)$.
            \STATE Compute Loss of $D_p$: $\mathcal{L}_{D} = \mathcal{L}_{CE}(D_p(f_p), a)$.
            \STATE Optimize $D_{p}$ using $\mathcal{L}_{D}$.
            \STATE Compute Fairness Loss of $G_p$: $\mathcal{L}_{G}^{fair} = -\mathcal{L}_{CE}(D_p(f_p), a) - \alpha \mathcal{H}(D_p(f_p))$.
            \STATE Compute Segmentation Loss of $G_p$: $\mathcal{L}_G^{seg} = \frac{1}{2}(\mathcal{L}_{CE}(y, \hat{y}) + \mathcal{L}_{Dice}(y, \hat{y}))$.
            \STATE Optimize $G_{p}$ using $\mathcal{L}_{G} = \mathcal{L}_G^{seg} + \beta\mathcal{L}_{G}^{fair}$.
        \ENDFOR
    \end{algorithmic}    
\end{algorithm}


\section{Experiments and Results}

\subsection{Experimental Configurations}

\noindent{\textbf{Dataset.}}
Two 2D medical segmentation datasets are used in this paper, including a Thyroid Ultrasound Cine-clip dataset~\cite{TUSC} (TUSC) consisting of 860 frames captured from ultrasound videos, and a skin lesion dataset~\cite{tschandl2018ham10000,codella2019skin} (ISIC 2018) composed by 10,015 RGB images.
Each dataset is split into a training set and a test set with a ratio of 7:3. We use \textit{sex} and \textit{age} as the sensitive attributes. Note that \textit{age} is categorized into 5 subgroups (0-20, 20-40, 40-60, 60-80, 80-100) to evaluate APPLE on unfairness mitigation with multi-class attributes.
The distribution of the two datasets is shown in the Supplementary.
We also applied two unfairness mitigation methods, re-sampling~\cite{puyol2021fairness} (RS) and subgroup models~\cite{puyol2021fairness} (SM) for performance comparison.

\noindent{\textbf{Fairness Metrics.}}
\label{fairmat}
Following~\cite{lee_investigation_2023}, we use maximum disparity ($\Delta$), skewed error rate (SER), and standard deviation (STD) as fairness metrics. Let $\mathcal{I}$ denote the segmentation utility, i.e. Dice, these fairness metrics can be computed as follows.

\begin{align}
    \mathcal{I}_{\Delta} = \max_k(\mathcal{I}^k) - \min_k(\mathcal{I}^k),~
    \mathcal{I}_{SER}    = \frac{\max_k(1 - \mathcal{I}^k)}{\min_k(1 - \mathcal{I}^k)},~
    \mathcal{I}_{STD}    = \sqrt{\frac{\sum_{k=1}^{K}(\mathcal{I}^k - \bar{\mathcal{I}})^2}{K-1}},
\end{align}
where $\bar{\mathcal{I}}=\frac{1}{K}\sum_{k=1}^K\mathcal{I}^k$, and $\mathcal{I}^k$ is the utility on the $k$-th subgroup.

\noindent{\textbf{Network Settings.}}
The images are resized to $256\times256$ and randomly flipped and rotated for data augmentation.
We first train baseline segmentors using U-Net~\cite{ronneberger2015u}, AttU-Net~\cite{wang2021attu}, and CMUNet~\cite{tang2023cmu} (U-Net only for ISIC 2018 dataset), the learning rate is set as $10^{-2}$, and SGD optimizer is used.
Then, we freeze the weights of these segmentors and train APPLE on them.
The learning rates of $G_p$ and $D_p$ are set as $10^{-3}$ and the Adam optimizer is adopted for model training.
The weighted factor $\alpha$ and $\beta$ are defined as 0.1 and 1.0, respectively.
All the experiments are conducted on a Ubuntu server with $8 \times$ NVIDIA 3090 Ti GPUs and repeated thrice.

\subsection{Main Results}

\noindent{\textbf{Results on TUSC Dataset.}}
The results of APPLE on the TUSC dataset are shown in Table~\ref{tab:main_exp_tusc}.
Compared to the baseline models, their APPLE-adopting counterpart models show better fairness scores on both binary \textit{sex} and multi-class \textit{age} attributes.
The disparity is more significant on \textit{age} (e.g. the Dice Disparity of CMUNet improves about $15\times10^{-2}$, from $17.06\times10^{-2}$ to $2.14\times10^{-2},$), which might due to the larger variance among subgroups.
Besides, although RS has better fairness on U-Net and CMUNet, it requires extra sensitive information in the inference. And it cannot tackle unfairness when the attribute is a multi-class variable.

\begin{table}[t]
    \centering
    \caption{Result on TUSC Dataset ($\text{Mean}_{\text{Std}}$). \textcolor{blue}{Best} and \textcolor{brown}{Second} in each pair are highlighted.}
    \label{tab:main_exp_tusc}
    \resizebox{\textwidth}{!}{
    \begin{tabular}{lrrrrrrrrr}
        \toprule
         \multirow{2}{*}{\textbf{Model}}  & \multicolumn{4}{c}{\textbf{Binary \textit{Sex}}} & &\multicolumn{4}{c}{\textbf{Multi-class \textit{Age}}} \\
        \cline{2-5}\cline{7-10}
                             &  \textbf{ Avg$\%\uparrow$} & \textbf{$\Delta\%\downarrow$} & \textbf{SER$\downarrow$} & \textbf{STD$\%\downarrow$} & & \textbf{Avg$\%\uparrow$} &\textbf{$\Delta\%\downarrow$} & \textbf{SER$\downarrow$} & \textbf{STD$\%\downarrow$}  \\
        \midrule
        U-Net~\cite{ronneberger2015u}        &   \textcolor{brown}{$77.30_{0.38}$} & $4.51_{2.54}$ & $1.21_{0.11}$ & $2.26_{1.27}$  &  &\textcolor{blue}{{$77.30_{0.38}$}} & $19.24_{3.80}$ & \textcolor{brown}{$2.42_{0.17}$} & $6.97_{1.23}$  \\
        \rowcolor{gray!10}
        U-Net + RS~\cite{puyol2021fairness} &  $74.31_{1.73}$ & $3.69_{0.14}$ & $1.15_{0.02}$ & $1.84_{0.07}$ & &\textcolor{brown}{$76.86_{2.16}$} & \textcolor{brown}{$15.39_{7.09}$} & {$3.03_{1.00}$} & \textcolor{brown}{$5.37_{2.24}$}  \\
        \rowcolor{gray!20}
        U-Net + SM~\cite{puyol2021fairness} &  \textcolor{blue}{{$79.49_{0.47}$}} & \textcolor{blue}{{$2.30_{0.07}$}} & \textcolor{blue}{{$1.12_{0.01}$}} & \textcolor{blue}{{$1.15_{0.04}$}} & & {$72.48_{0.24}$}  & {$56.91_{0.02}$} & {$4.24_{0.04}$} & {$24.07_{0.12}$} \\
        \rowcolor{gray!30}
        U-Net + APPLE &  {$74.18_{1.25}$} & \textcolor{brown}{$3.54_{1.85}$} & \textcolor{brown}{$1.15_{0.07}$} & \textcolor{brown}{$1.77_{0.92}$} & & {$73.35_{0.67}$}&\textcolor{blue}{$6.75_{0.60}$} & \textcolor{blue}{$1.31_{0.03}$} & \textcolor{blue}{$2.37_{0.10}$} \\
        \midrule
        AttU-Net~\cite{wang2021attu}        &  \textcolor{brown}{$77.70_{0.60}$} & $3.70_{2.41}$ & $1.18_{0.11}$ & $1.85_{1.21}$ & &\textcolor{blue}{$77.70_{0.60}$}& $20.34_{4.62}$ & $2.41_{0.16}$ & $7.73_{2.10}$\\
        \rowcolor{gray!10}
        AttU-Net + RS~\cite{puyol2021fairness} &  {$73.79_{0.71}$} & {$2.67_{0.33}$}  & \textcolor{brown}{$1.11_{0.01}$} & \textcolor{brown}{$1.33_{0.16}$} && \textcolor{brown}{$75.65_{1.43}$} & \textcolor{brown}{$11.15_{3.25}$} & \textcolor{brown}{$1.69_{0.25}$} & \textcolor{brown}{$4.04_{1.17}$} \\
        \rowcolor{gray!20}
        AttU-Net + SM~\cite{puyol2021fairness} &  \textcolor{blue}{$80.42_{0.40}$} & \textcolor{brown}{$2.66_{0.83}$} & {$1.15_{0.05}$} & {$1.33_{0.41}$} & &{$74.11_{0.24}$} & {$58.74_{0.60}$} & {$4.79_{0.16}$}& {$24.03_{0.13}$} \\
        \rowcolor{gray!30}
        AttU-Net + APPLE &  $72.90_{0.27}$ & \textcolor{blue}{$2.06_{2.04}$} & \textcolor{blue}{$1.09_{0.09}$} & \textcolor{blue}{$1.03_{1.02}$} & & {$73.07_{0.26}$} & \textcolor{blue}{$8.94_{0.73}$} & \textcolor{blue}{$1.44_{0.05}$} & \textcolor{blue}{$3.59_{0.30}$}  \\
        \midrule					
        CMUNet~\cite{tang2023cmu}        &  $82.93_{0.95}$ & $3.97_{0.78}$ & $1.28_{0.08}$ & $1.99_{0.39}$ &  & $82.93_{0.95}$ &\textcolor{brown}{$17.06_{10.10}$} & \textcolor{brown}{$1.66_{0.52}$} & \textcolor{brown}{$6.20_{3.76}$} \\
        \rowcolor{gray!10}
        CMUNet + RS~\cite{puyol2021fairness} &  {$80.20_{1.79}$} & {$3.83_{1.07}$} & {$1.21_{0.03}$} & {$1.92_{0.54}$}  && \textcolor{blue}{$84.72_{1.23}$} & {$17.57_{9.24}$} & {$3.03_{1.00}$} & {$6.53_{3.51}$}\\
        \rowcolor{gray!20}
        CMUNet + SM~\cite{puyol2021fairness} &  \textcolor{blue}{$85.47_{0.15}$} & \textcolor{blue}{$1.77_{0.34}$}  & \textcolor{blue}{$1.13_{0.03}$} & \textcolor{blue}{$0.89_{0.17}$}  && {$74.64_{0.19}$} & {$59.15_{0.17}$} & {$4.92_{0.07}$} & {$27.08_{0.12}$}\\
        \rowcolor{gray!30}
        CMUNet + APPLE &  \textcolor{brown}{$83.00_{1.07}$} & \textcolor{brown}{$2.81_{0.56}$} & \textcolor{brown}{$1.19_{0.06}$} & \textcolor{brown}{$1.41_{0.28}$}  && \textcolor{brown}{$83.45_{0.19}$}& \textcolor{blue}{$2.14_{0.62}$} & \textcolor{blue}{$1.15_{0.05}$} & \textcolor{blue}{$0.93_{0.27}$} \\
        \bottomrule
    \end{tabular}
    }
\end{table}

\noindent{\textbf{Results on ISIC 2018 Dataset.}}
As shown in Table~\ref{tab:main_exp_isic}, APPLE occupies the best fairness scores on \textit{sex} among the four algorithms. This is followed by RS, which has almost all the Second utility and fairness.
Similar results can be found on multi-class \textit{age}, except for STD, which does not outperform the base model.
SM still increases fairness gaps on \textit{age}, which might be due to the insufficient information learned by subgroup models.

\begin{table}[t]
    \centering
    \caption{Result on ISIC2018 Dataset ($\text{Mean}_{\text{Std}}$). \textcolor{blue}{Best} and \textcolor{brown}{Second} in each pair are highlighted.}
    \label{tab:main_exp_isic}
    \resizebox{\textwidth}{!}{
    \begin{tabular}{lrrrrrrrrr}
        \toprule
         \multirow{2}{*}{\textbf{Model}}  & \multicolumn{4}{c}{\textbf{Binary \textit{Sex}}} & &\multicolumn{4}{c}{\textbf{Mult-class \textit{Age}}} \\
        \cline{2-5}\cline{7-10}
                             &   \textbf{Avg$\%\uparrow$} & \textbf{$\Delta\%\downarrow$} & \textbf{SER$\downarrow$} & \textbf{STD$\%\downarrow$} & & \textbf{Avg$\%\uparrow$} &\textbf{$\Delta\%\downarrow$} & \textbf{SER$\downarrow$} & \textbf{STD$\%\downarrow$}  \\
        \midrule
        U-Net~\cite{ronneberger2015u}         &  \textcolor{blue}{$91.32_{0.02}$} & {$0.58_{0.01}$} & {$1.07_{0.01}$} & {$0.29_{0.01}$}   && {$91.43_{0.21}$} & \textcolor{brown}{$5.89_{0.12}$} & \textcolor{brown}{$1.92_{0.01}$} & \textcolor{blue}{$2.11_{0.03}$}      \\
       \rowcolor{gray!10}
        U-Net + RS~\cite{puyol2021fairness} & \textcolor{brown}{$90.84_{0.13}$} & \textcolor{brown}{$0.24_{0.04}$} & \textcolor{brown}{$1.03_{0.01}$} & \textcolor{brown}{$0.12_{0.02}$}  &&  \textcolor{brown}{$91.57_{0.03}$} & {$5.92_{0.11}$} & {$1.96_{0.03}$} & {$2.25_{0.01}$}  \\
        \rowcolor{gray!20}
        U-Net + SM~\cite{puyol2021fairness} &  {$90.03_{0.03}$} & {$1.21_{0.02}$} & {$1.13_{0.01}$} & {$0.61_{0.00}$} &&  {$86.81_{0.30}$} & {$20.14_{6.29}$} & {$5.46_{0.67}$} & {$17.11_{1.64}$} \\
        \rowcolor{gray!30}
        U-Net + APPLE &  {$89.89_{0.35}$} & \textcolor{blue}{$0.11_{0.10}$} & \textcolor{blue}{$1.01_{0.01}$} & \textcolor{blue}{$0.06_{0.05}$}  &&  \textcolor{blue}{$91.59_{0.04}$} & \textcolor{blue}{$5.71_{0.07}$} & \textcolor{blue}{$1.90_{0.04}$} & \textcolor{brown}{$2.20_{0.02}$} \\
        \bottomrule
    \end{tabular}
    }
\end{table}

\subsection{Ablation Study}
Ablation studies are conducted on the \textit{sex} attribute of the TUSC dataset.
First, we change weight factor $\beta$ of $L_{G}^{fair}$ to 0.1 and 5.0. The result is shown in Table.~\ref{tab:ablation_hyper}. As $\beta$ is used to control the weight of $L_{G}^{fair}$, a higher $\beta$ should result in lower $\Delta$, SER, and STD, which is proved by the first part of Table.~\ref{tab:ablation_hyper}.
Although $\beta=5.0$ obtains the best fairness scores, the significant drop in utility ($\sim17\%$) is unacceptable in real applications. Thus, it is important to choose a proper $\beta$ to balance the utility and fairness.

Besides, we also integrate APPLE with RS to evaluate whether APPLE can further improve model fairness. As shown in the second part of Table~\ref{tab:ablation_hyper}, APPLE can improve both utility and fairness compared to RS, which shows good scalability with other mitigation methods.

\begin{table}[t]
    \centering
    \caption{Ablation Study ($\text{Mean}_{\text{Std}}$). \textcolor{blue}{Best} and \textcolor{brown}{Second} in each pair are highlighted.}
    \label{tab:ablation_hyper}
    \resizebox{0.8\textwidth}{!}{
    \begin{tabular}{lrrrr}
        \toprule
        \multicolumn{5}{c}{\textbf{Ablation on Hyper-parameter $\beta$}} \\
        \midrule
        {\textbf{Model}} &  \textbf{Avg$\%\uparrow$} & \textbf{$\Delta\%\downarrow$} & \textbf{SER$\downarrow$} & \textbf{STD$\%\downarrow$} \\
        \midrule
        U-Net~\cite{ronneberger2015u}        &   \textcolor{blue}{$77.30_{0.38}$} & $4.51_{2.54}$ & $1.21_{0.11}$ & $2.26_{1.27}$  \\
        \midrule
        \rowcolor{gray!10}
        U-Net + APPLE ($\beta=0.1$) &  $73.85_{1.63}$ & $4.68_{1.64}$ & $1.19_{0.05}$ & $2.34_{0.82}$ \\
        \midrule
        \rowcolor{gray!20}
        U-Net + APPLE ($\beta=1.0$) &   \textcolor{brown}{$74.18_{1.25}$} & \textcolor{brown}{$3.54_{1.85}$} & \textcolor{brown}{$1.15_{0.07}$} & \textcolor{brown}{$1.77_{0.92}$} \\
        \midrule
        \rowcolor{gray!30}
        U-Net + APPLE ($\beta=5.0$) &  $60.08_{14.17}$ &  \textcolor{blue}{$2.44_{1.85}$} &  \textcolor{blue}{$1.06_{0.03}$} &  \textcolor{blue}{$1.22_{0.93}$} \\
        \midrule
        \multicolumn{5}{c}{\textbf{Scalability with Other Mitigation Method}} \\
        \midrule
        U-Net + RS~\cite{puyol2021fairness}        &  \textcolor{brown}{$74.31_{1.73}$} & \textcolor{brown}{$3.69_{0.14}$} & \textcolor{brown}{$1.15_{0.02}$} & \textcolor{brown}{$1.84_{0.07}$} \\
        \midrule
        \rowcolor{gray!15}
        U-Net + RS + APPLE  &  \textcolor{blue}{$76.39_{0.35}$} &\textcolor{blue}{ $2.57_{0.10}$} & \textcolor{blue}{$1.12_{0.01}$} & \textcolor{blue}{$1.29_{0.05}$}  \\
        \bottomrule
    \end{tabular}
    }
\end{table}

Moreover, we also evaluate whether APPLE can be integrated into foundation models, including SAM~\cite{kirillov2023segany} and MedSAM~\cite{MedSAM}. 
The experiments are conducted on the ISIC 2018 dataset and \textit{sex} is chosen as the sensitive attribute. We train SAM + APPLE and MedSAM + APPLE using the official pre-trained weights and pipeline.
The box prompt is defined as the bounding box of the ground truth segmentation mask.
As illustrated in Table~\ref{tab:sam}, our APPLE shows promising fairness on the two pre-trained models. Moreover, although applying APPLE to SAM decreases the overall utility by about $7\%$, the overall performance increases when adopted to MedSAM ($91.64\% \rightarrow 92.68\%$), which provides a potential for mitigating unfairness in foundation models in real medical applications for improving healthy equity.

\begin{table}[t]
    \centering
    \caption{Integrability on two large-scale foundation models ($\text{Mean}_{\text{Std}}$). \textcolor{blue}{Better} in each pair are highlighted.}
    \label{tab:sam}
    \resizebox{\textwidth}{!}{
    \begin{tabular}{lrrrrrrrrr}
        \toprule
         \multirow{2}{*}{\textbf{Model}}  & \multicolumn{4}{c}{\textbf{SAM}~\cite{kirillov2023segany}} & &\multicolumn{4}{c}{\textbf{MedSAM}~\cite{MedSAM}} \\
        \cline{2-5}\cline{7-10}
                             &   \textbf{Avg$\%\uparrow$} & \textbf{$\Delta\%\downarrow$ }& \textbf{SER$\downarrow$} & \textbf{STD$\%\downarrow$} & & \textbf{Avg$\%\uparrow$} &\textbf{$\Delta\%\downarrow$} & \textbf{SER$\downarrow$} & \textbf{STD$\%\downarrow$}  \\
        \midrule
        Baseline     & \textcolor{blue}{$83.40_{0.08}$} & {$1.51_{0.08}$} & {$1.10_{0.00}$} & {$0.75_{0.04}$}  && {$91.64_{0.03}$} & {$0.73_{0.08}$} & {$1.09_{0.01}$} & {$0.36_{0.04}$}    \\
       \rowcolor{gray!15}
        Baseline + APPLE &  {$76.03_{0.56}$} & \textcolor{blue}{$0.55_{0.13}$} & \textcolor{blue}{$1.02_{0.01}$} & \textcolor{blue}{$0.27_{0.07}$} &&  \textcolor{blue}{$92.58_{0.01}$} & \textcolor{blue}{$0.56_{0.04}$} & \textcolor{blue}{$1.08_{0.01}$} & \textcolor{blue}{$0.28_{0.02}$} \\
        \bottomrule
    \end{tabular}
    }
    
\end{table}

\section{Conclusion}
In this paper we propose a novel algorithm, termed APPLE, to improve model fairness without modifying the architectures or parameters of the pre-trained base models in medical segmentation tasks.
This is achieved by perturbing the latent embedding of the base model using a GAN, which aims to manipulate the embedding such that no sensitive information can be passed to the decoder of the segmentor.
Extensive experiments on two medical segmentation datasets prove the effectiveness of APPLE on three base segmentors.
By conducting ablation studies, APPLE presents its promising ability to tackle unfairness issues cooperated with other unfairness mitigation methods and foundation models.
Further research will be conducted on validating the utility of APPLE on other tasks including detection and reconstruction, and large-scale vision-language models such as CARzero\cite{lai2024carzero} to protect health equity.

\bibliographystyle{splncs04}
\bibliography{mybib}

\begin{thebibliography}{10}
\providecommand{\url}[1]{\texttt{#1}}
\providecommand{\urlprefix}{URL }
\providecommand{\doi}[1]{https://doi.org/#1}

\bibitem{bissoto2020debiasing}
Bissoto, A., Valle, E., Avila, S.: Debiasing skin lesion datasets and models? not so fast. In: Proceedings of the IEEE/CVF Conference on Computer Vision and Pattern Recognition Workshops. pp. 740--741 (2020)

\bibitem{booth2021bias}
Booth, B.M., Hickman, L., Subburaj, S.K., Tay, L., Woo, S.E., D'Mello, S.K.: Bias and fairness in multimodal machine learning: A case study of automated video interviews. In: Proceedings of the 2021 International Conference on Multimodal Interaction. pp. 268--277 (2021)

\bibitem{burlina2021addressing}
Burlina, P., Joshi, N., Paul, W., Pacheco, K.D., Bressler, N.M.: Addressing artificial intelligence bias in retinal diagnostics. Translational Vision Science \& Technology  \textbf{10}(2),  13--13 (2021)

\bibitem{codella2019skin}
Codella, N., Rotemberg, V., Tschandl, P., Celebi, M.E., Dusza, S., Gutman, D., Helba, B., Kalloo, A., Liopyris, K., Marchetti, M., et~al.: Skin lesion analysis toward melanoma detection 2018: A challenge hosted by the international skin imaging collaboration (isic). arXiv preprint arXiv:1902.03368  (2019)

\bibitem{TUSC}
shared datasets, S.A.: Thyroid ultrasound cine-clip. \url{https://stanfordaimi.azurewebsites.net/datasets/a72f2b02-7b53-4c5d-963c-d7253220bfd5}

\bibitem{deng2023fairness}
Deng, W., Zhong, Y., Dou, Q., Li, X.: On fairness of medical image classification with multiple sensitive attributes via learning orthogonal representations. In: International Conference on Information Processing in Medical Imaging. pp. 158--169. Springer (2023)

\bibitem{du2023unveiling}
Du, Y., Xue, Y., Dharmakumar, R., Tsaftaris, S.A.: Unveiling fairness biases in deep learning-based brain mri reconstruction. In: Workshop on Clinical Image-Based Procedures. pp. 102--111. Springer (2023)

\bibitem{dwork2012fairness}
Dwork, C., Hardt, M., Pitassi, T., Reingold, O., Zemel, R.: Fairness through awareness. In: Proceedings of the 3rd innovations in theoretical computer science conference. pp. 214--226 (2012)

\bibitem{hardt2016equality}
Hardt, M., Price, E., Srebro, N.: Equality of opportunity in supervised learning. Advances in neural information processing systems  \textbf{29} (2016)

\bibitem{ioannou2022study}
Ioannou, S., Chockler, H., Hammers, A., King, A.P., Initiative, A.D.N.: A study of demographic bias in cnn-based brain mr segmentation. In: MLCN 2022, Held in Conjunction with MICCAI 2022, Singapore, September 18, 2022, Proceedings. pp. 13--22. Springer (2022)

\bibitem{kim2019learning}
Kim, B., Kim, H., Kim, K., Kim, S., Kim, J.: Learning not to learn: Training deep neural networks with biased data. In: Proceedings of the IEEE/CVF conference on computer vision and pattern recognition. pp. 9012--9020 (2019)

\bibitem{kirillov2023segany}
Kirillov, A., Mintun, E., Ravi, N., Mao, H., Rolland, C., Gustafson, L., Xiao, T., Whitehead, S., Berg, A.C., Lo, W.Y., Doll{\'a}r, P., Girshick, R.: Segment anything. arXiv:2304.02643  (2023)

\bibitem{lai2024carzero}
Lai, H., Yao, Q., Jiang, Z., Wang, R., He, Z., Tao, X., Zhou, S.K.: Carzero: Cross-attention alignment for radiology zero-shot classification. arXiv preprint arXiv:2402.17417  (2024)

\bibitem{lee_investigation_2023}
Lee, T., Puyol-Antón, E., Ruijsink, B., Aitcheson, K., Shi, M., King, A.P.: An investigation into the impact of deep learning model choice on sex and race bias in cardiac {MR} segmentation (Aug 2023), \url{http://arxiv.org/abs/2308.13415}, arXiv:2308.13415 [cs, eess]

\bibitem{MedSAM}
Ma, J., He, Y., Li, F., Han, L., You, C., Wang, B.: Segment anything in medical images. Nature Communications  \textbf{15}, ~1--9 (2024)

\bibitem{pakzad2022circle}
Pakzad, A., Abhishek, K., Hamarneh, G.: Circle: Color invariant representation learning for unbiased classification of skin lesions. In: European Conference on Computer Vision. pp. 203--219. Springer (2022)

\bibitem{petersen2022feature}
Petersen, E., Feragen, A., Zemsch, L.d.C., Henriksen, A., Christensen, O.E.W., Ganz, M.: Feature robustness and sex differences in medical imaging: a case study in mri-based alzheimer's disease detection. arXiv preprint arXiv:2204.01737  (2022)

\bibitem{puyol2021fairness}
Puyol-Ant{\'o}n, E., Ruijsink, B., Piechnik, S.K., Neubauer, S., Petersen, S.E., Razavi, R., King, A.P.: Fairness in cardiac mr image analysis: an investigation of bias due to data imbalance in deep learning based segmentation. In: MICCAI 2021. pp. 413--423. Springer (2021)

\bibitem{quan2024slide}
Quan, Q., Tang, F., Xu, Z., Zhu, H., Zhou, S.K.: Slide-sam: Medical sam meets sliding window. In: Medical Imaging with Deep Learning (2024)

\bibitem{ronneberger2015u}
Ronneberger, O., Fischer, P., Brox, T.: U-net: Convolutional networks for biomedical image segmentation. In: MICCAI 2015. pp. 234--241. Springer (2015)

\bibitem{seyyed2021underdiagnosis}
Seyyed-Kalantari, L., Zhang, H., McDermott, M.B., Chen, I.Y., Ghassemi, M.: Underdiagnosis bias of artificial intelligence algorithms applied to chest radiographs in under-served patient populations. Nature medicine  \textbf{27}(12),  2176--2182 (2021)

\bibitem{tang2023cmu}
Tang, F., Wang, L., Ning, C., Xian, M., Ding, J.: Cmu-net: a strong convmixer-based medical ultrasound image segmentation network. In: ISBI 2023. pp.~1--5. IEEE (2023)

\bibitem{tschandl2018ham10000}
Tschandl, P., Rosendahl, C., Kittler, H.: The ham10000 dataset, a large collection of multi-source dermatoscopic images of common pigmented skin lesions. Scientific data  \textbf{5}(1), ~1--9 (2018)

\bibitem{wang2021attu}
Wang, S., Li, L., Zhuang, X.: Attu-net: attention u-net for brain tumor segmentation. In: International MICCAI Brainlesion Workshop. pp. 302--311. Springer (2021)

\bibitem{wang2022fairness}
Wang, Z., Dong, X., Xue, H., Zhang, Z., Chiu, W., Wei, T., Ren, K.: Fairness-aware adversarial perturbation towards bias mitigation for deployed deep models. In: Proc. IEEE Comput. Vis. Patten Recog. pp. 10379--10388 (2022)

\bibitem{wu2022fairprune}
Wu, Y., Zeng, D., Xu, X., Shi, Y., Hu, J.: Fairprune: Achieving fairness through pruning for dermatological disease diagnosis. In: MICCAI 2023. pp. 743--753. Springer (2022)

\bibitem{xiao2018generating}
Xiao, C., Li, B., Zhu, J.Y., He, W., Liu, M., Song, D.: Generating adversarial examples with adversarial networks. arXiv preprint arXiv:1801.02610  (2018)

\bibitem{xu2023fairness}
Xu, Z., Li, J., Yao, Q., Li, H., Zhou, S.K.: Fairness in medical image analysis and healthcare: A literature survey. Authorea Preprints  (2023)

\bibitem{xu2023fairadabn}
Xu, Z., Zhao, S., Quan, Q., Yao, Q., Zhou, S.K.: Fairadabn: Mitigating unfairness with adaptive batch normalization and its application to dermatological disease classification. arXiv preprint arXiv:2303.08325  (2023)

\bibitem{yao2022improving}
Yao, R., Cui, Z., Li, X., Gu, L.: Improving fairness in image classification via sketching. arXiv preprint arXiv:2211.00168  (2022)

\bibitem{pmlr-v174-zhang22a}
Zhang, H., Dullerud, N., Roth, K., Oakden-Rayner, L., Pfohl, S., Ghassemi, M.: Improving the fairness of chest x-ray classifiers. In: Flores, G., Chen, G.H., Pollard, T., Ho, J.C., Naumann, T. (eds.) Proceedings of the Conference on Health, Inference, and Learning. Proceedings of Machine Learning Research, vol.~174, pp. 204--233. PMLR (07--08 Apr 2022), \url{https://proceedings.mlr.press/v174/zhang22a.html}

\bibitem{zhou2021review}
Zhou, S.K., Greenspan, H., Davatzikos, C., Duncan, J.S., Van~Ginneken, B., Madabhushi, A., Prince, J.L., Rueckert, D., Summers, R.M.: A review of deep learning in medical imaging: Imaging traits, technology trends, case studies with progress highlights, and future promises. Proceedings of the IEEE  (2021)

\end{thebibliography}

\end{document}